\documentclass[journal]{IEEEtran}
\usepackage{booktabs}
\usepackage[pdftex]{graphicx}
\usepackage{longtable}
\usepackage{subfigure}
\usepackage{graphicx}

\ifCLASSINFOpdf
\else
\fi
\begin{document}
%
\title{Feature Selection Convolutional Neural Networks for Visual Tracking}
%
%
%

\author{Zhiyan~Cui,
        Na~Lu*
        
}

\maketitle

\begin{abstract}
Most of the existing tracking methods based on CNN(convolutional neural networks) are too slow for real-time application despite the excellent tracking precision compared with the traditional ones. Moreover, neural networks are memory intensive which will take up lots of hardware resources. In this paper, a feature selection visual tracking algorithm combining CNN based MDNet(Multi-Domain Network) and RoIAlign was developed. We find that there is a lot of redundancy in feature maps from convolutional layers. So valid feature maps are selected by mutual information and others are abandoned which can reduce the complexity and computation of the network and do not affect the precision. The major problem of MDNet also lies in the time efficiency. Considering the computational complexity of MDNet is mainly caused by the large amount of convolution operations and fine-tuning of the network during tracking, a RoIAlign layer which could conduct the convolution over the whole image instead of each RoI is added to accelerate the convolution and a new strategy of fine-tuning the fully-connected layers is used to accelerate the update. With RoIAlign employed, the computation speed has been increased and it shows greater precision than RoIPool. Because RoIAlign can process float number coordinates by bilinear interpolation. These strategies can accelerate the processing, reduce the complexity with very low impact on precision and it can run at around 10 fps(while the speed of MDNet is about 1 fps). The proposed algorithm has been evaluated on a benchmark: OTB100, on which high precision and speed have been obtained.

\end{abstract}

\begin{IEEEkeywords}
visual tracking, RoIAlign, feature selection, convolutional neural networks
\end{IEEEkeywords}

%
\IEEEpeerreviewmaketitle

\section{Introduction}
\IEEEPARstart{V}{isual} tracking is one of the fundamental problems in computer vision which aims at estimating the position of a predefined target in an image sequence, with only its initial state given. Nowadays, CNN has achieved great success in computer vision, such as object classification~\cite{22,30}, object detection~\cite{10,9,28}, semantic segmentation~\cite{10,14,11} and so on. However, it is difficult for CNN to play a great role in visual tracking. CNN usually consists of a large amount of parameters and needs big dataset to train the network to avoid over-fitting. So far, there are some novel studies~\cite{23,5,27,34} which combine CNN and traditional trackers to achieve the start of the art. Some of them just use CNNs which are trained on ImageNet or other large datasets to extract features.\\
On one hand, tracking seems much easier than detection because it only needs to conduct a binary classification between the target and the background. On the other hand, it is difficult to train the network because of the diversity of objects that we might track. An object may be the target in one video but the background in another. And there is no such amount of data for tracking to train a deep network.\\
MDNet~\cite{26} is a novel solution for tracking problem with detection method. It is trained by learning the shared representations of targets from multiple video sequences, where each video is regarded as a separate domain. There are several branches in the last layer of the network for binary classification and each of them is corresponding to a video sequence. The preceding layers are in charge of capturing common representations of targets. While tracking, the last layer is removed and a new single branch is added to the network to compute the scores in the test sequences. Parameters in the fully-connected layers are fine-tuned online during tracking. Both long and short updating strategies are conducted for robustness and adaptability, respectively. Hard negative mining~\cite{32} technique is involved in the learning procedure. MDNet won the first place in VOT2015 competition. Although the effect is amazing, MDNet runs at very low speed which is 1 fps on GPU. \\
In this paper, we propose a much faster network with less computation for object tracking, referred to as Feature Selection Network(FSNet). It consists of 3 convolution layers and 3 fully-connected layers with a RoIAlign layer~\cite{14} between them. The convolution is operated on the whole image instead of each RoI, and a RoIAlign layer is used to get fixed-size features which can enhance the calculation speed. When we compare the results of RoIAlign and RoIPool~\cite{9}, it shows that RoIAlign could obtain higher precision. Instead of fine-tuning the network with fixed number of iterations, we use a threshold to adjust the numbers dynamically, which will reduce the iterations and time of fine-tuning. The channel of feature maps from the third convolutional layer is 512 and nearly half of them are redundant. So we just select 256 useful feature maps by mutual information. This feature map selection can reduce nearly half of the parameters in the network and do not affect the precision.\\
Similar training and multi-domain strategy as in MDNet have been adopted. Given $k$ videos to train the network, the last fully-connected layer should have $k$ branches. Each video is corresponding to one branch. So the last layer is domain-specific layers and others are shared layers. In this way, the shared information of the tracked objects is captured. The $k$ branches of last layer are deleted and a new one is added to the network before tracking. The parameters in last layer are initialized randomly. The positive and negative samples which are extracted from the first frame of the sequence are used to fine-tune the network. Only parameters in the fully-connected layers will be updated. Bounding boxes in the current frame around the previous target position are sent to the network for discrimination. The sample with the highest score will be obtained as the target in this frame. The fully-connected layers will be updated with samples from the previous several frames when scores of all the samples are less than a threshold. By this dynamic method, the network can learn the new features of the objects.\\
The proposed algorithm consists of multi-domain learning, feature map selection and dynamic network tracking. The main contributions are shown below:\\
\begin{itemize}
	\item A small neural network with RoIAlign layer is created to accelerate the convolution. 
	\item A novel feature map selection method is proposed to reduce the complexity and the computation of the network and do not have any influence on the precision.
	\item A new strategy of fine-tuning the fully-connected layers is used to reduce the iterations during tracking. About 10 fps on GPU could be reached which makes it  faster than MDNet with only a small decrease in precision.
	\item The performance of different structures of the network has been compared. The result shows that RoIAlign outperforms RoIPool which is usually used in detection and classification. And neither increasing convolution layers nor decreasing fully-connected layers has positive effects on the accuracy. 
	\item Evaluations of the proposed tracker have been performed on a public benchmark: object tracking benchmark~\cite{36}. The results show that a higher precision and faster tracking speed than most of the existing trackers based on CNN have been obtained.\\	
	
	
\end{itemize}
\section{Related Work}
There have been several surveys~\cite{39,38} of visual tracking during these years. Most of the state of the art methods are based on correlation filter or deep learning. Some trackers have been combined with each other to improve the tracking performance.\\
Correlation filter (CF) has attracted scientists' attention for several years. It can run at high speed because of the application of Fourier transformation. The calculation is transformed from spatial domain to Fourier domain to speed up the operation. MOSSE~\cite{4} was the first algorithm which uses CF to track objects. CSK~\cite{16} is another algorithm based on CF. It uses circular shift to produce dense samples by making use of more features of the image. And then, there are KCF~\cite{17} and DCF~\cite{17}. Both of them can use multi-channel features of images. DSST~\cite{6} uses translation filter and scale filter to deal with changes in location and scale. Because of Fourier transformation, most of these trackers can reach a very high speed.\\ 
Nowadays, deep learning has been widely used in computer vision, such as classification, detection and semantic segmentation. Many trackers based on deep learning have been designed to get a better precision. Some of them combine deep learning with CF. DLT~\cite{37} is the first tracker to use deep learning. The structure is based on particle filter and it uses SDAE to extract features. HCF~\cite{23} uses the features in different layers from a neural network and then applies CF to conduct tracking. MDNet~\cite{26} is a novel tracker based on CNN which uses detection method to realize tracking. It samples bounding box around the target location in the next frame and sends them to the network to find the one with the highest score, which is the target in the next frame. And the network will be fine-tuned when it does not work well any longer. TCNN~\cite{25} is a tracker based on CNN and a tree structure. It builds a tree to evaluate the reliability of the model. GOTURN~\cite{15} is the fastest tracker based on CNN and it can achieve 165 fps. It uses the deep neural network to regress the bounding box in the next frame. But its target localization precision is poor. In a word, the high precision of a tracker based on CNN is obtained at the sacrifice of speed.

\begin{figure}[htp]
	\begin{center}
		\includegraphics[width=0.45\textwidth]{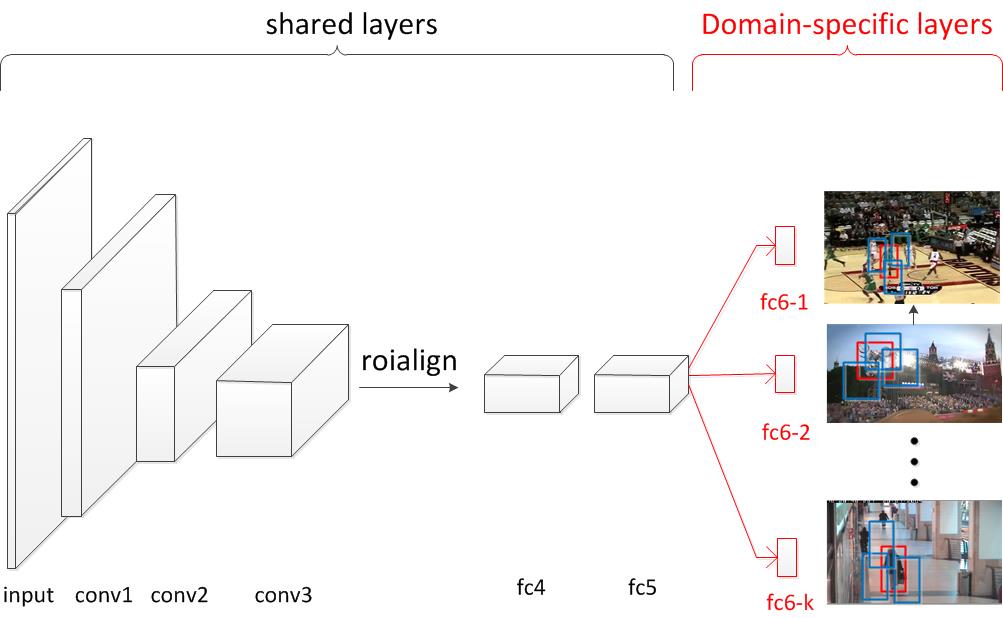}
		\caption{The architecture of fast dynamic convolutional neural network. Red and blue bounding boxes denote the positive and negative samples, respectively.}\label{fig:network}
	\end{center}
\end{figure}

\begin{figure}[htp]
	\begin{center}
		\includegraphics[width=0.4\textwidth]{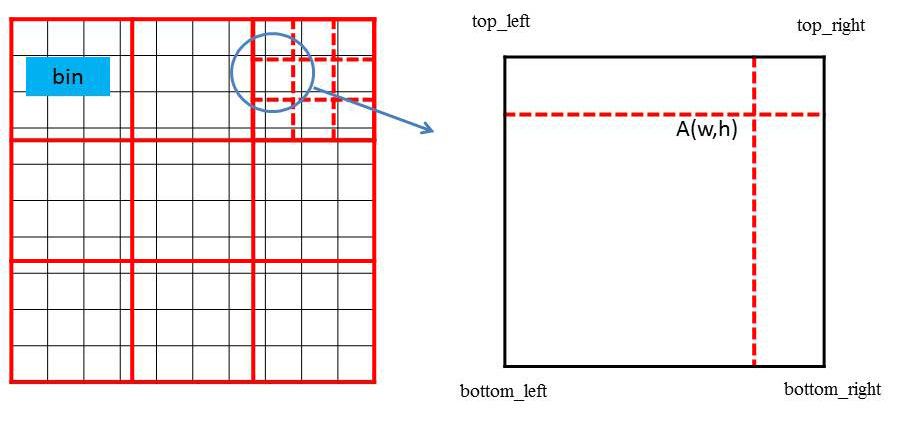}
		\caption{The schematic diagram of RoIAlign. Suppose there is a 10x10 pixels RoI (in black) and we want to get a 3x3 feature (in red). 2x2 points are selected in each bin. Because the locations of the points are float types, we use four integer points nearby and bilinear interpolation to get the value of this point. And the biggest one is used to represent this bin (max RoIAlign).}\label{fig:align}
	\end{center}
\end{figure}

\begin{figure*} 	
	\centering 	
	\includegraphics[width=0.10\linewidth]{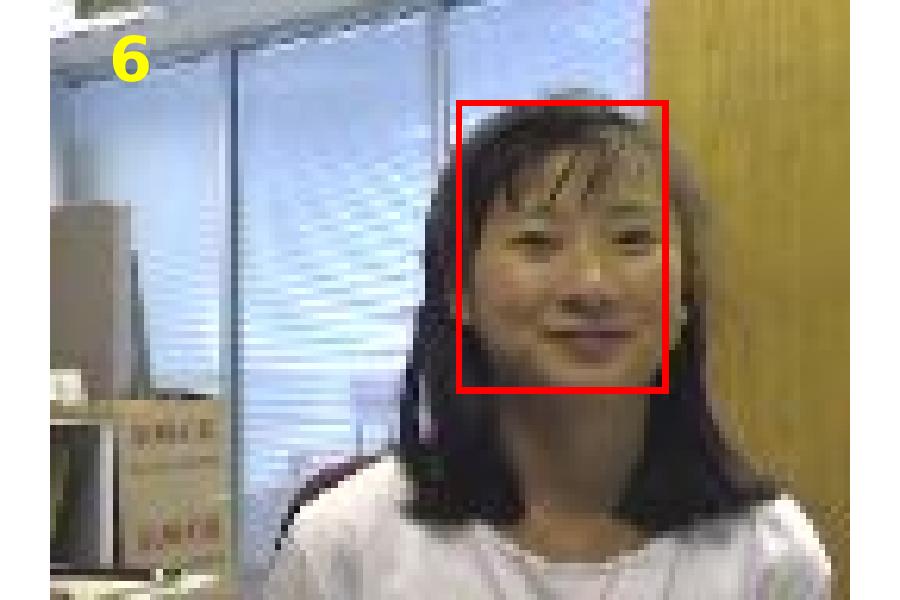}
	\includegraphics[width=0.85\linewidth]{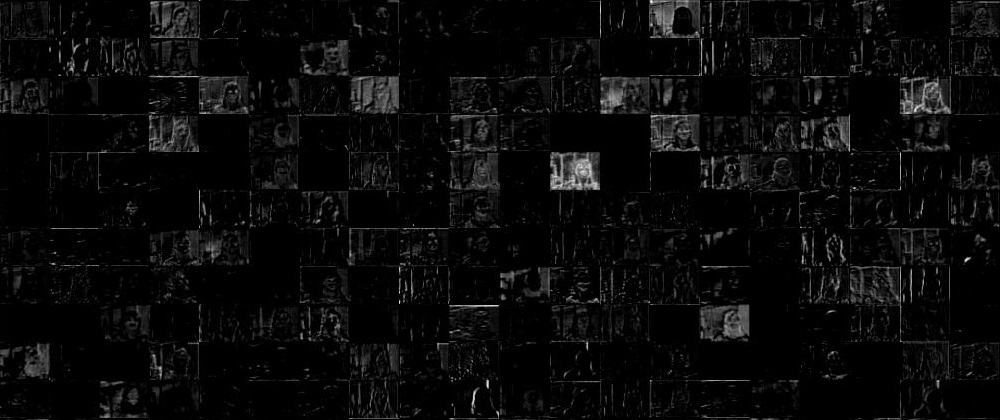}	
	\caption{An example of the feature map visualization from conv3. The left shows the original image in OTB100 dataset. And the right shows corresponding feature maps(after ReLU layer). There are 512 feature maps in conv3 and only some of them are shown.}
	\label{fig:visualize}	
\end{figure*}
 \begin{figure}[htp]
	\begin{center}
		\includegraphics[width=0.45\textwidth]{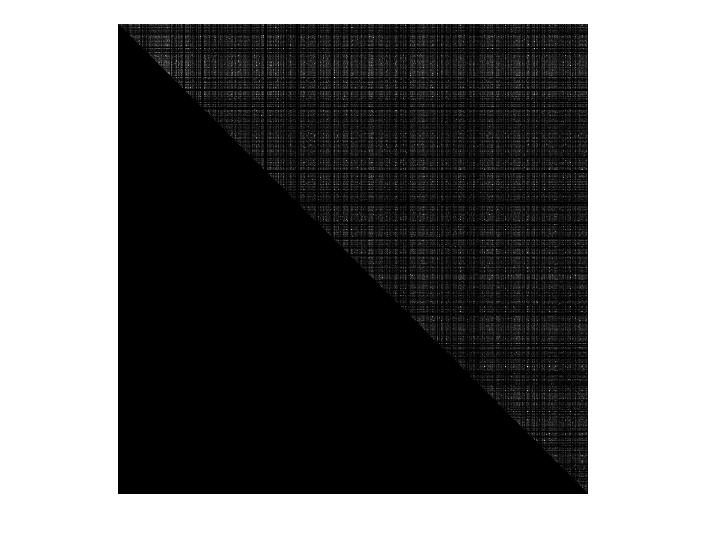}
		\caption{The mutual information for all pairs of feature maps. The brighter the pixel, the higher mutual information they have. For clarity, we only show the results of some feature maps. And the mutual information of a feature map itself is defined as 0. So the pixels on diagonal are all pure black. We know that the matrix for mutual information is a symmetric matrix. So just half of the values are illustrated and others are set to zero.}\label{fig:mutual_info}
	\end{center}
\end{figure}

\begin{figure*} 	
	\centering 	
	\includegraphics[width=0.20\linewidth]{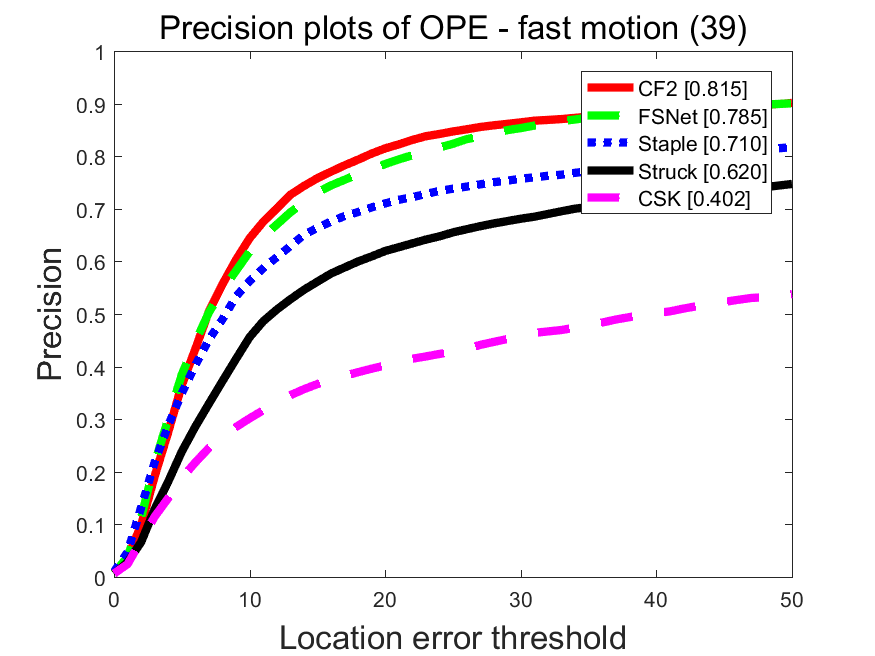}
	\includegraphics[width=0.20\linewidth]{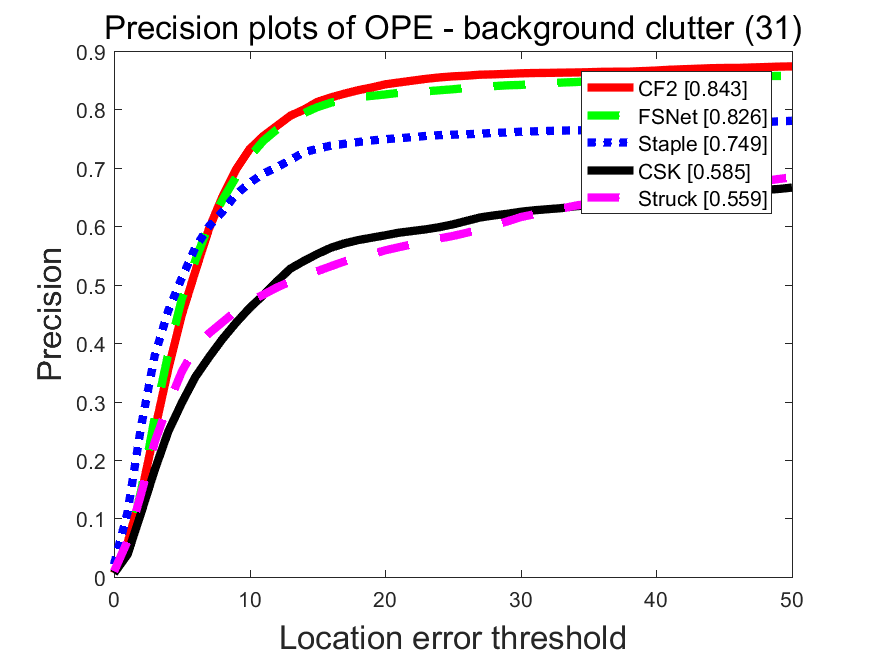}
	\includegraphics[width=0.20\linewidth]{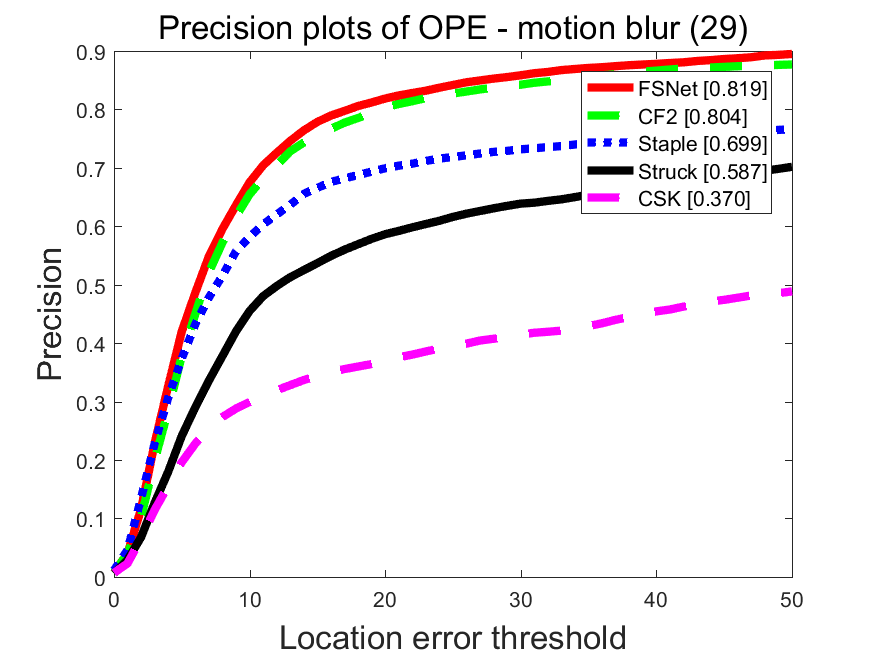}
	\includegraphics[width=0.20\linewidth]{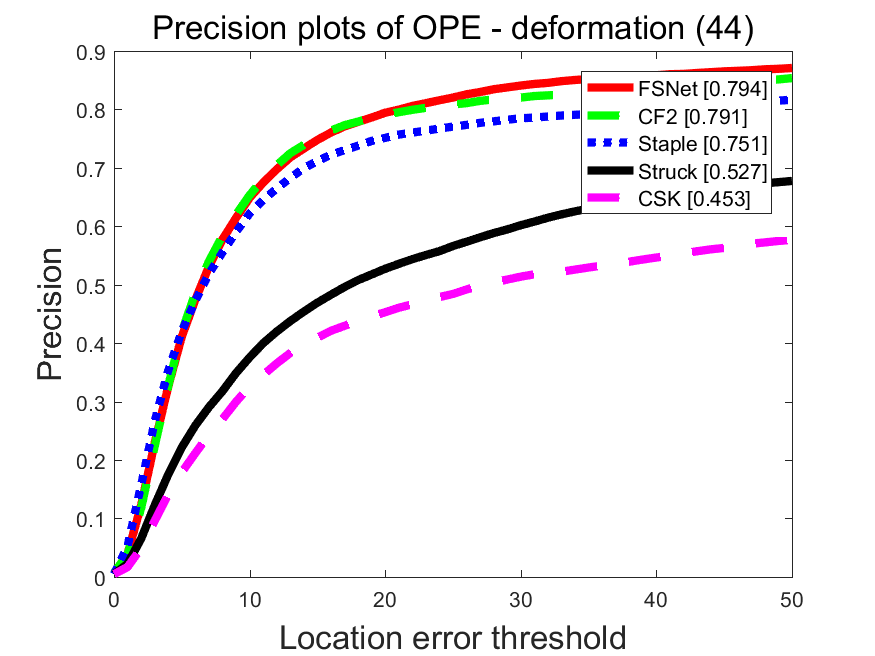}   

	\includegraphics[width=0.20\linewidth]{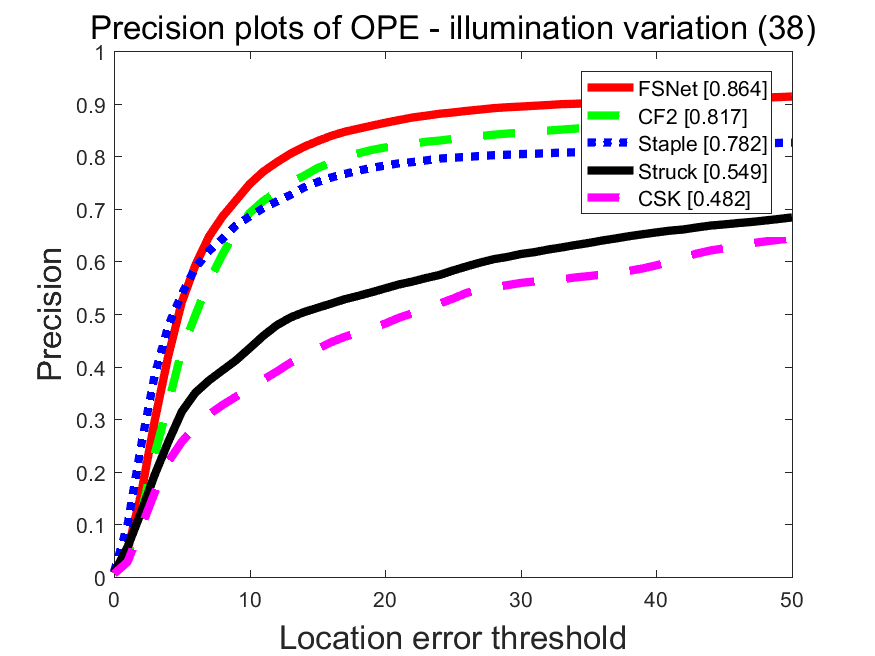}
	\includegraphics[width=0.20\linewidth]{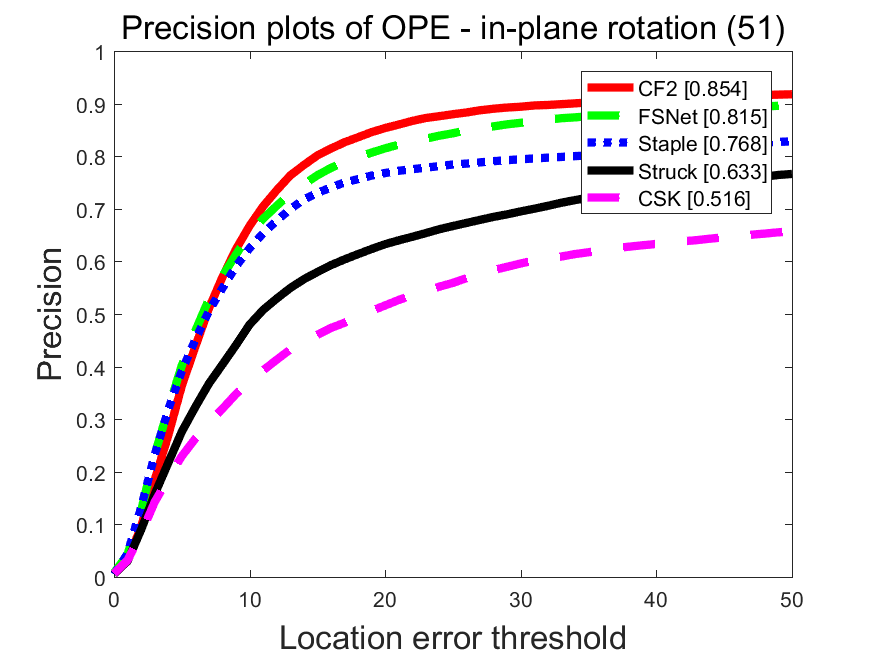}
	\includegraphics[width=0.20\linewidth]{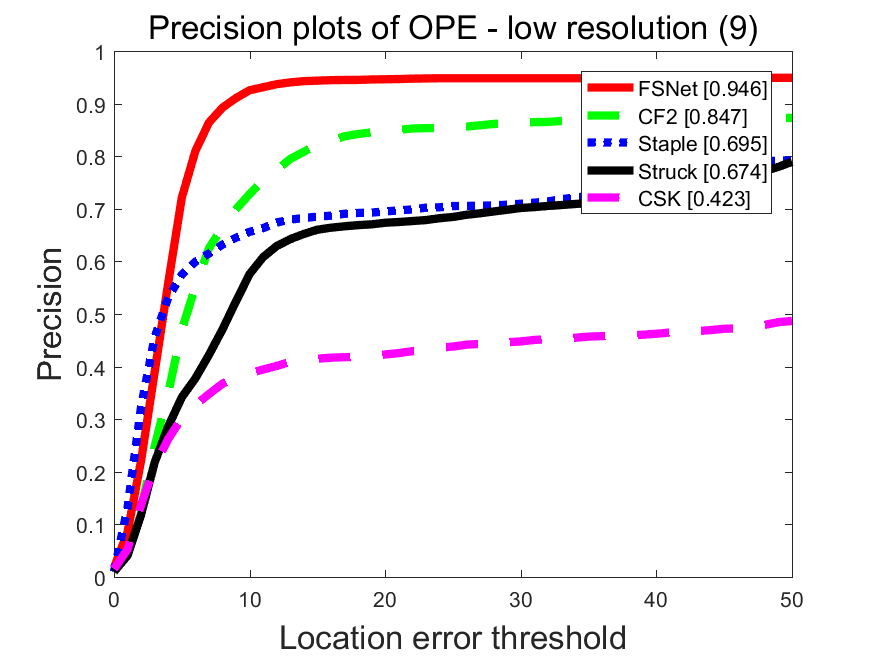}
	\includegraphics[width=0.20\linewidth]{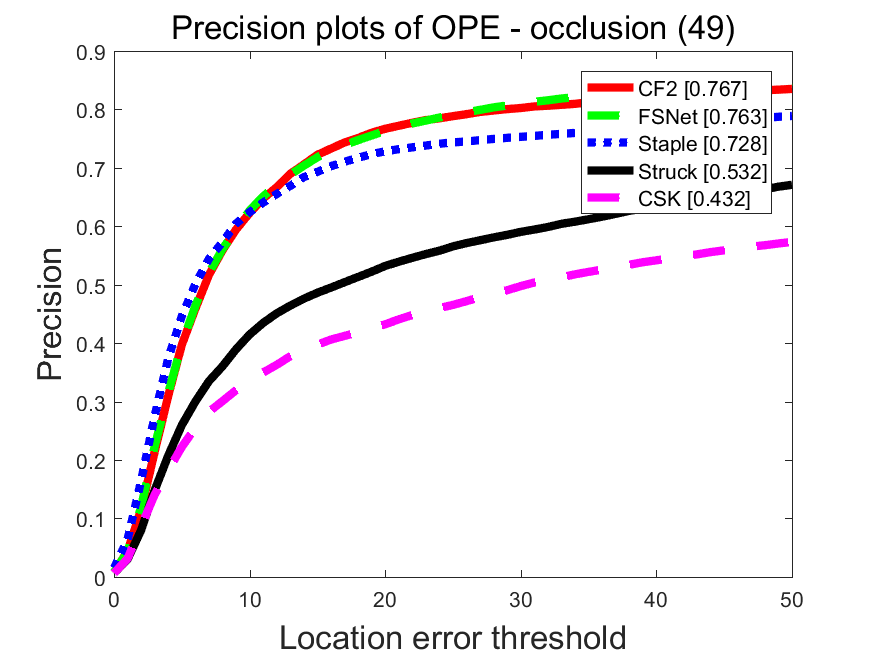}  
	
	\includegraphics[width=0.20\linewidth]{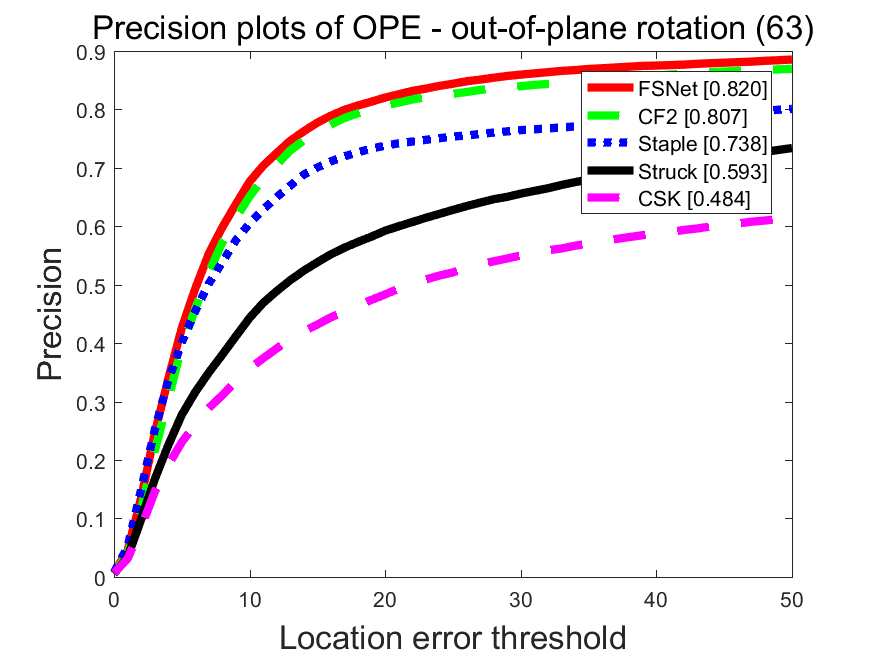}
	\includegraphics[width=0.20\linewidth]{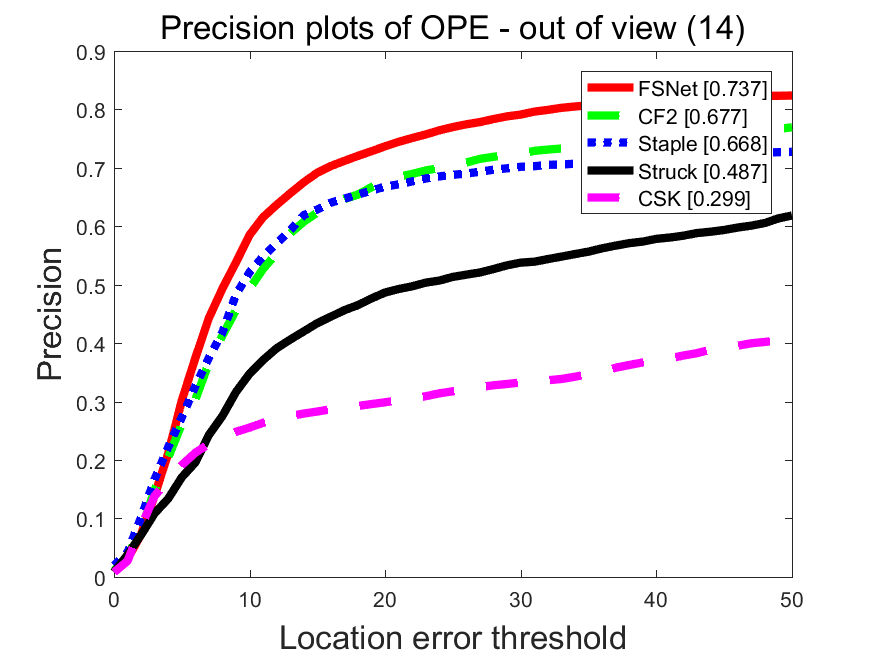}
	\includegraphics[width=0.20\linewidth]{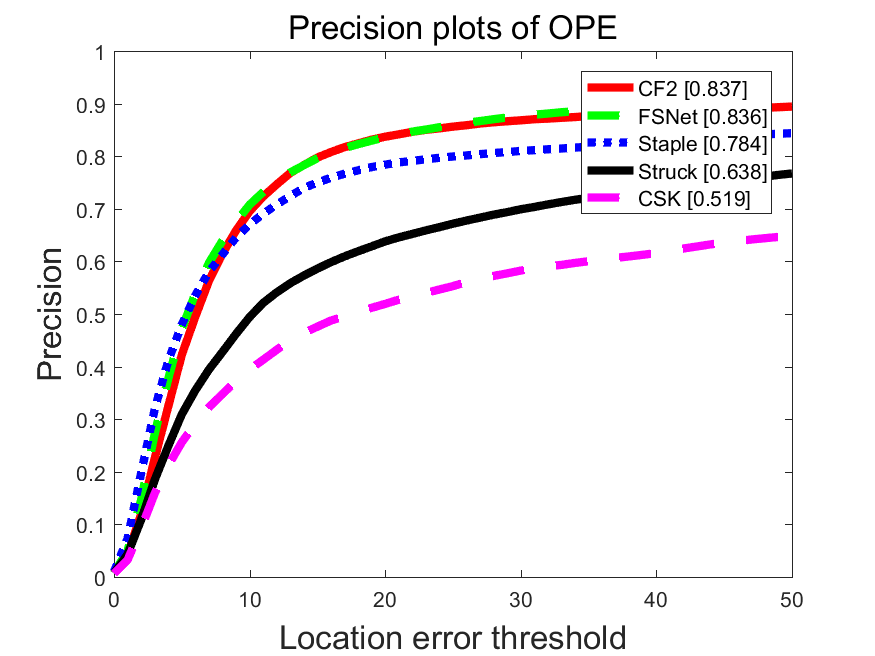}
	\includegraphics[width=0.20\linewidth]{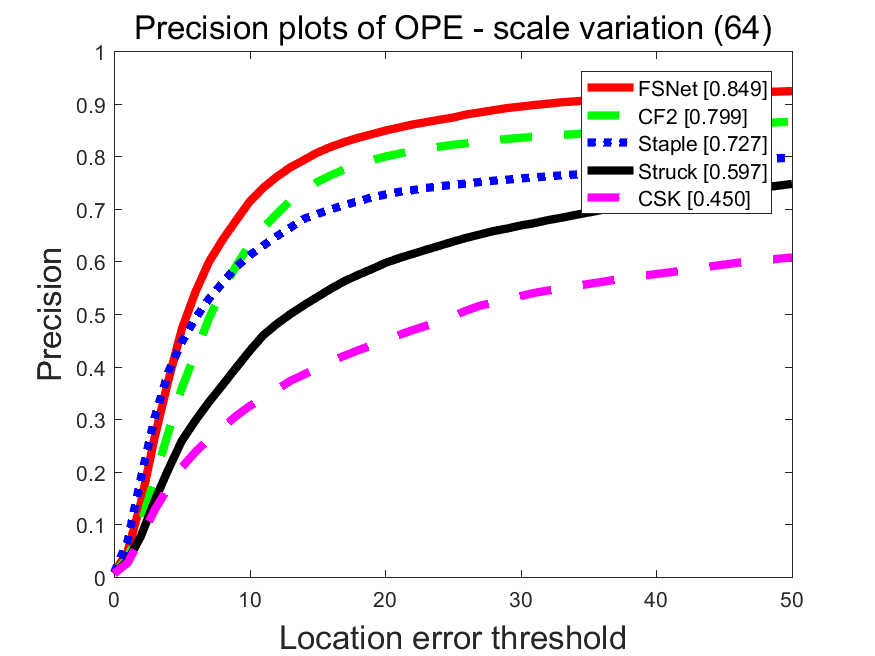}   
	
	
	\caption{Results on OTB100}
	\label{fig:OTB100}	
\end{figure*}

\begin{figure*} 	
	\centering 	
	
	\includegraphics[width=0.20\linewidth]{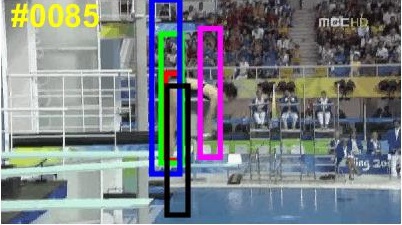}
	\includegraphics[width=0.20\linewidth]{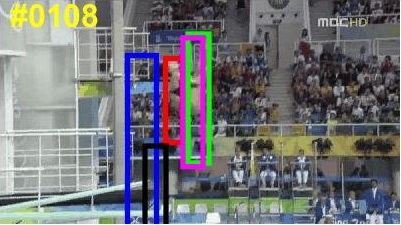}
	\includegraphics[width=0.20\linewidth]{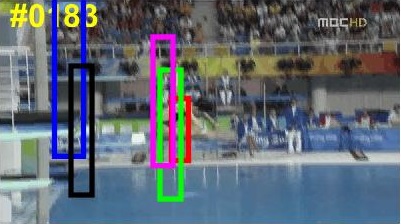}
	\includegraphics[width=0.20\linewidth]{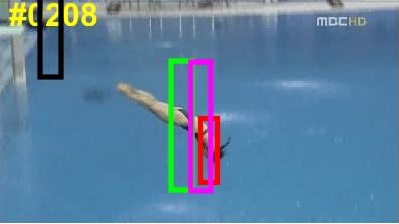}

	\includegraphics[width=0.20\linewidth]{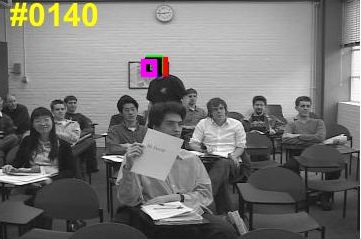}
	\includegraphics[width=0.20\linewidth]{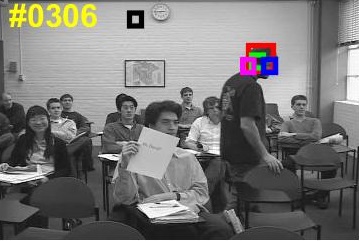}
	\includegraphics[width=0.20\linewidth]{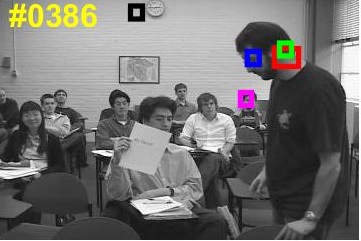}
	\includegraphics[width=0.20\linewidth]{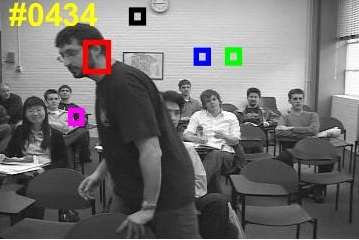}

	\includegraphics[width=0.20\linewidth]{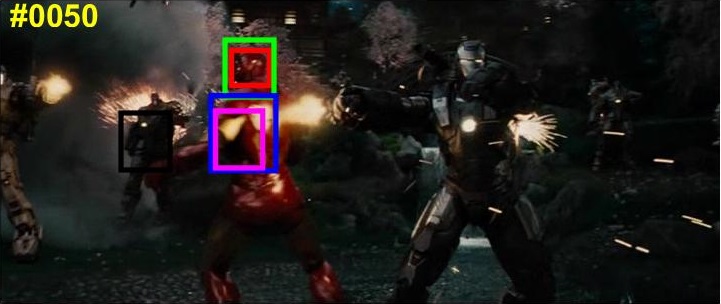}
	\includegraphics[width=0.20\linewidth]{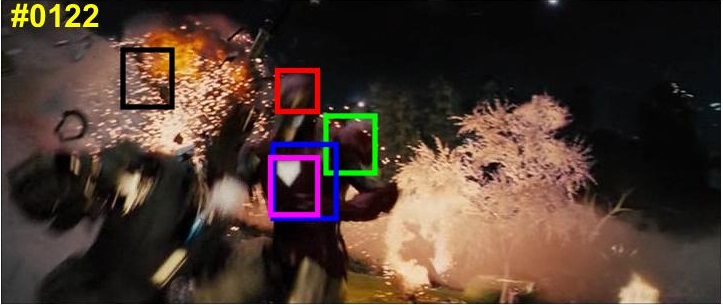}
	\includegraphics[width=0.20\linewidth]{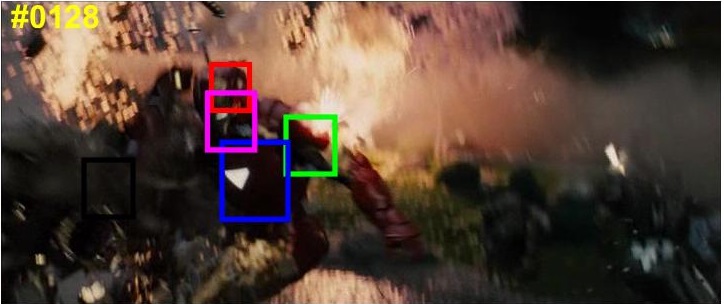}
	\includegraphics[width=0.20\linewidth]{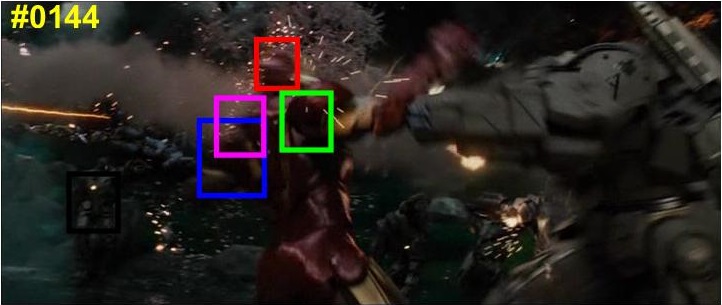}   
	
	\includegraphics[width=0.20\linewidth]{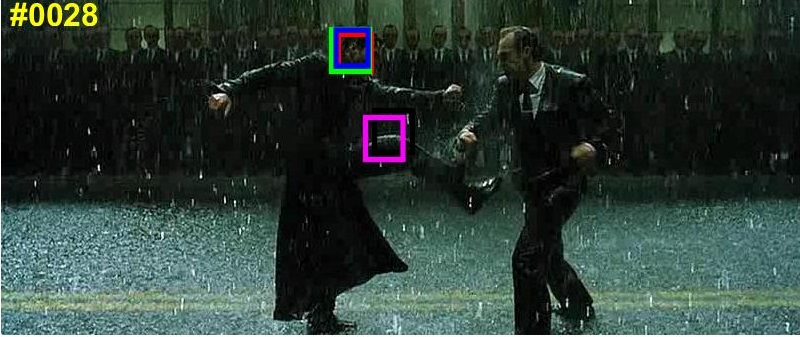}
	\includegraphics[width=0.20\linewidth]{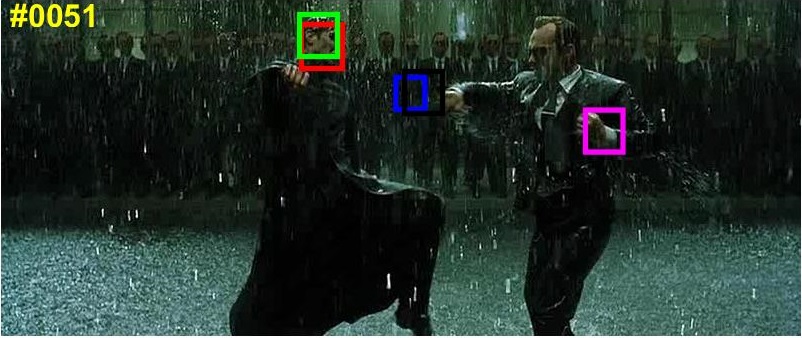}
	\includegraphics[width=0.20\linewidth]{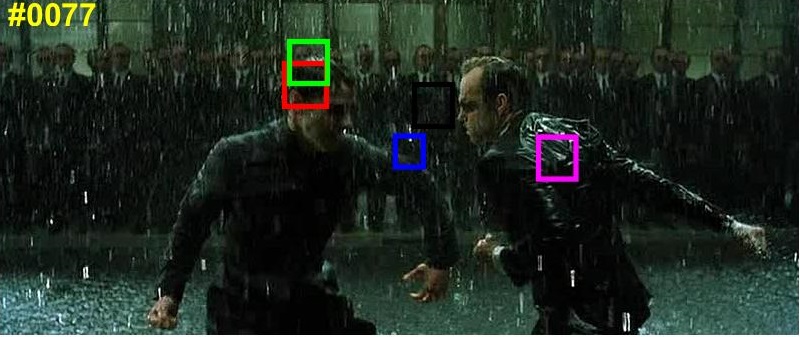}
	\includegraphics[width=0.20\linewidth]{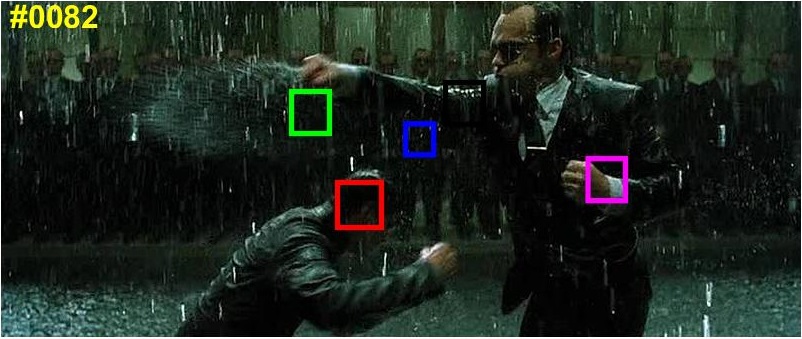}  
	
	\includegraphics[width=0.20\linewidth]{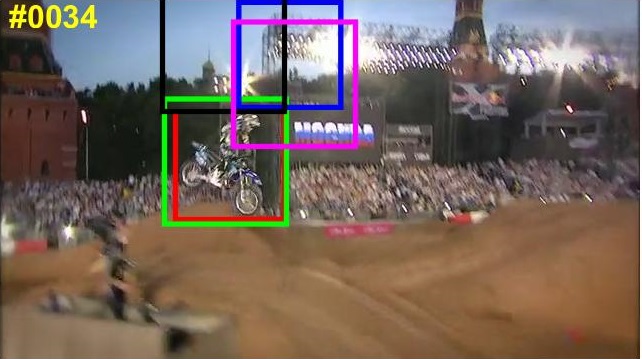}
	\includegraphics[width=0.20\linewidth]{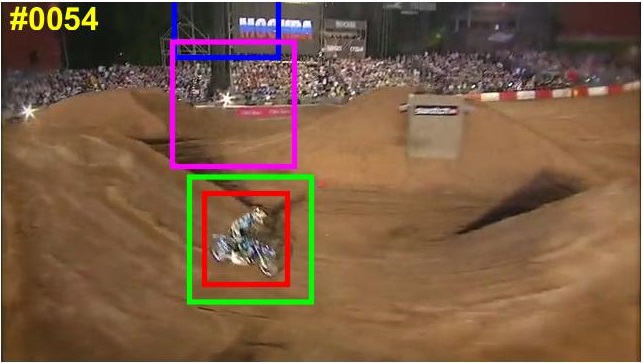}
	\includegraphics[width=0.20\linewidth]{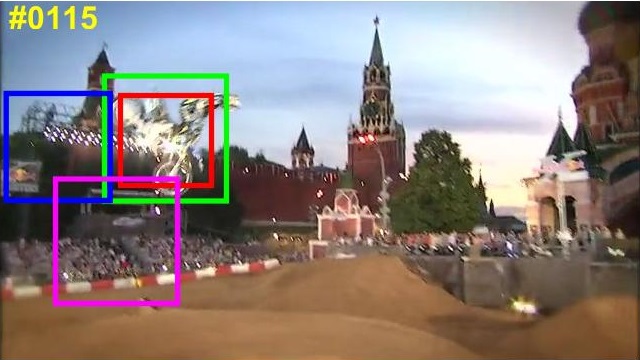}
	\includegraphics[width=0.20\linewidth]{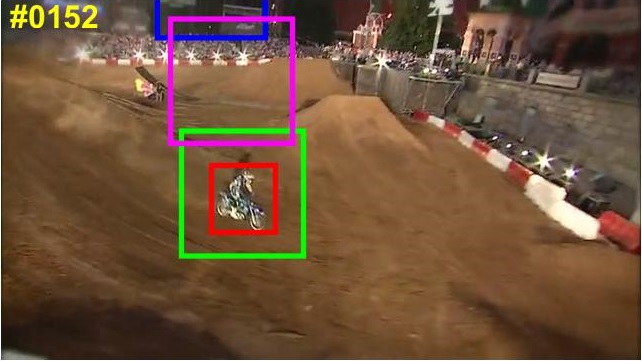}   
	
	\includegraphics[width=0.70\linewidth]{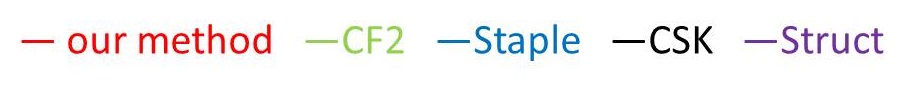}   
	
	\caption{Comparison of some visual tracking examples}
	\label{fig:OTBresults}	
\end{figure*}

\section{Feature Selection Convolutional Neural Networks}
This section describes the structure of the network and how to train and test the proposed network. The convolution is operated on the whole image and RoIAlign is used to obtain fix-sized features. This strategy can speed up the convolution computation and maintain the localization accuracy in the meanwhile. Different branches have been added to the last layer in correspondence to different video sequences, and the preceding layers are shared layers to get common representations of targets through training. During the tracking stage, only half of the feature maps from the last convolutional layers are used and  the last fully-connected layer is removed and a new one with only one branch is added to the network. Parameters in the fully-connected layers will be updated in a new strategy which will reduce the ineffective updates during tracking to adapt to the currently tracked targets.
\subsection{Convolution on the Whole Image}
In MDNet, there are 256 RoIs extracted from each frame and all of them are scaled to a fixed size. Then they will be sent to the convolutional layers and get fixed-size features which can then be delivered to the fully-connected layers. But it is a waste of time to do so. It contains a lot of redundancy calculation and the convolution operation has to repeat for many times which will definitely increase the time of process.\\
To speed up the convolution, a strategy of conducting convolution on the whole image is adopted and it only needs to do the convolutional operation for once. At the end of the convolution layers, a RoIPool or RoIAlign is used to extracted fixed-size features corresponding to the RoIs from the last convolution layer. The experiments show that RoIAlign is better than RoIPool in precision which we will discuss the details in section 4.
\subsection{Network Architecture}
The architecture of our network is illustrated in Fig.~\ref{fig:network}. The first three convolution layers are borrowed from the VGG-M~\cite{45} network. The size of the filters in the first convolution layer is $7\times7\times3\times96$, which is followed by a Relu, a normalization~\cite{19} and a pooling layer. The size of the filters in the second convolution layer is $5\times5\times96\times256$, also followed by a Relu, a normalization and a pooling layer. The size in the third convolution layer is $3\times3\times256\times512$, followed by a Relu layer. Then there is a RoIAlign layer to extract fixed size features ( here we set it to $3\times3\times512$) from the convolution layers. Next, there are two fully-connected layers with dropout ( the rate is set to 0.5). The last fully-connected layer is just like the one in MDNet. Suppose we use $k$ videos to train the network, the network will have $k$ branches in the last layer and each of them uses two labels to represent the target and the background. The last layer is domain-specific layer and the others are shared layers. That means, when the network is trained with the $i\textsuperscript{th}(i=1,2,…,k)$ video, we just update the parameters in $i\textsuperscript{th}$ branch of the last fully-connected layer and the shared layers.

Most of the neural networks for classification and detection use Pool or RoIPool layer after the convolution layers. RoIPool can get fixed-size (e.g., $3\times3$) features from each RoI, which can then be sent to the following fully-connected layers. However, there is one major issue with RoIPool. After several convolution operations, the size and position of the RoIs might be float numbers, and we need to divide the RoIs into fixed-size regions (e.g., $3\times3$). The RoIPool rounds the float numbers to the nearest integers to fulfill the pooling. The localization precision may get lost in this operation.\\
RoIAlign~\cite{14} is another way to get fixed-size(e.g., $3\times3$) features from each RoI. It can work better than RoIPool in theory. Meanwhile, RoIAlign does not spend much more time than RoIPool. The performance comparison of RoIPool and RoIAlign could be found in section 4. Different from RoIPool, RoIAlign keeps the float numbers in the operation. At the last step of RoIAlign, we use bilinear interpolation~\cite{46} to calculate $n$ points in each bin and use the largest one to represent this bin(max-RoIAlign). \\
Fig.~\ref{fig:align} shows an example of RoIAlign. The point $A(w,h)$ is what we need and its coordinates are float. The four points:$top\_left(x_{l},y_{t})$,$top\_right(x_{r},y_{t})$,$bottom\_left(x_{l},y_{b})$ and $bottom\_right(x_{r},y_{b})$ around it are the nearest integer points. And the values of these points are $Value_{t\_l}$, $Value_{t\_r}$, $Value_{b\_l}$ and $Value_{b\_r}$. The value of $A$ could be calculated via the following equations:\\
\\
\hspace*{0.1cm}$V_1=(1-(w-x_{l}))\times(1-(h-y_{t}))\times Value_{t\_l}$\\
\hspace*{0.1cm}$V_2=(1-(x_{r}-w))\times(1-(h-y_{t}))\times Value_{t\_r}$\\
\hspace*{0.1cm}$V_3=(1-(w-x_{l}))\times(1-(y_{b}-h))\times Value_{b\_l}$\\
\hspace*{0.1cm}$V_4=(1-(x_{r}-w))\times(1-(y_{b-h}))\times Value_{b\_r}$\\
\hspace*{0.1cm}$Value_A=V_1+V_2+V_3+V_4$

When it comes to back propagation, just like RoIPool, we just use the points which contribute to the network and propagate their errors. Suppose that point $A$ is contributed to the network and its gradient is represented by $Value_{der}$ , and the back propagation is like that:\\
\hspace*{0.1cm}$Value_{t\_l}=(1-(w-\left\lfloor w\right\rfloor))\times (1-(h-\left\lfloor h\right\rfloor))\times Value_{der}$\\
\hspace*{0.1cm}$Value_{t\_r}=(1-(w-\left\lfloor w\right\rfloor))\times (1-(\left\lceil h\right\rceil-h))\times Value_{der}$\\
\hspace*{0.1cm}$Value_{b\_l}=(1-(\left\lceil w\right\rceil-w))\times (1-(h-\left\lfloor h\right\rfloor))\times Value_{der}$\\
\hspace*{0.1cm}$Value_{b\_r}=(1-(\left\lceil w\right\rceil-w))\times (1-(\left\lceil h\right\rceil-h))\times Value_{der}$\\
For each bin, $2\times2$ points are chosen and we use the largest one to represent this bin. The precision has dropped slightly if $1\times1$ point is used and $3\times3$ points which will increase the calculation indicates that there is almost no difference with $2\times2$. So $2\times2$ points in each bin is an acceptable choice.\\
A smaller network than those which usually used in the detection or classification is constructed in our study. The most serious issue of the deep network solution for image tracking is the low time efficiency. Considering that in the tracking scenario only binary classification is involved, so a small and fast network can satisfy the need of tracking task. Otherwise, with more convolution layers added, the localization information will get diminished over layers which would influence the tracking accuracy. Some other network structures are compared with our network and the results show the superiority of the proposed network. The details could be found in section 4.
\subsection{Training Method}
Softmax cross-entropy loss and SGD are used to update the network. When $i\textsuperscript{th}$ video is used to train the network, $i\textsuperscript{th}$ branch of the last layer and shared layers are working together and their parameters are updated at the same time. Other branches in the last layer are disabled and do not work. It was shown in ~\cite{26} that $k$ branches could get better performance than only one branch in the last layer during training.\\
Samples are extracted around the target from each frame in a video. The samples whose IoU are larger than a threshold $t1$ are used as positive examples, and the ones with IoU less than a threshold $t2$ are used as negative examples. In our experiment, we set $t1$ to 0.7 and $t2$ to 0.5.\\
The three convolution layers have the same structure as in VGG-M network which have been pre-trained on ImageNet. The parameters in the fully-connected layers are initialized randomly which subject to Gaussian distribution. Because the convolution layers have been trained on ImageNet, a smaller learning rate is adopted from the beginning. We set the learning rate to 0.0001 and weight decay to 0.0005. It takes 100 iterations to get the local optimum.
\subsection{Feature map Selection before Tracking}
We proposed a feature map selection method to select useful features and reduce the computation. Our experiments show that it has no influence on the precision. There are 512 feature maps in the third convolutional layer(the last covolutional layer). In order to have an intuitive understand, an example of feature map visualization are shown in Fig.~\ref{fig:visualize}. We can conclude from Fig.~\ref{fig:visualize} that, first of all, some feature maps are almost no activation on the whole image. Secondly, it might be a lot of redundancy in these feature maps. Because tracking a target only needs to identify two categories, the target and background( some networks with 512 feature maps in the last convolutional layers can classify 1000 objects, e.g. VGGNet). Meanwhile, more than 90\% parameters of the network are in the fully-connected layer, and most of the parameters in fully-connected layers are in the first fully-connected layer. Because the parameters in first fully-connected layer has to connect all the neurons in the last convolutional layer and the first fully-connected layer. A lot of complexity and computation can be reduced if we can abandon some feature maps.\\

Considering these situation, we propose a feature map selection method based on mutual information to select useful features from the third convolutional layer. And our experiments illustrate that it has no influence on the precision. Formally, the mutual information of two discrete random varialbes $X$ and $Y$ can be defined as:\\
$$I(X,Y)=\sum_{x\in X}^{}\sum_{y\in Y}^{}p(x,y)log(\frac{p(x,y)}{p(x)p(y)})$$,\\
where $p(a,b)$ is the joint probability distribution of $X$ and $Y$, and $p(a)$ and $p(b)$ are their marginal probability distributions, respectively.
The mutual information between two feature maps is estimated using histograms of them. The activation values are distributed in 20 bins and the we can get probabilities for each bin. The mutual information of any two feature maps are calculated and these values are stored in a 512x512 matrix as show in Fig.~\ref{fig:mutual_info}( The pixel values are normalized. The matrix should be a symmetric matrix, but we only need half of it). There is a very interesting phenomenon from this matrix. We can see lots of black and white striped lines in the figure. The white line means this feature map often has lots of mutual information with all the others. On the contrary, the black line means this feature map usually has little mutual information with others. So we can see that there are a lot of redundancy in white lines and these corresponding feature maps can be abandoned. So at the beginning, feature maps whose values are all zero are deleted. Then, We will calculate mutual information of one feature map between others and find the maximum to represent this feature map(It means try to find a maximum in a row or a column in this matrix). Then sort these values and keep the smaller 256 feature maps.\\
Why should we select feature maps before tracking? Can we just use 256 feature maps in con3 when the network is trained? We show the results of training the network with 256 feature maps in section 4 and the precision will drop a little. The reason for this is that we use many kind of videos to train the network and different feature maps are activated by different objects. For examples, if we want to track a girl just like in Fig.~\ref{fig:visualize}, the feature maps which will activated by trees, animals or vehicles are useless for our purpose and there are almost no activation in these feature maps. But if we want to track another object, these useless feature maps when tracking a girl will be activated and became useful. We can not get enough features for all kinds of object if we just use 256 feature maps to train the network.\\

\subsection{Tracking Method}
\begin{figure} 	
	\centering 	
	\includegraphics[width=0.24\linewidth]{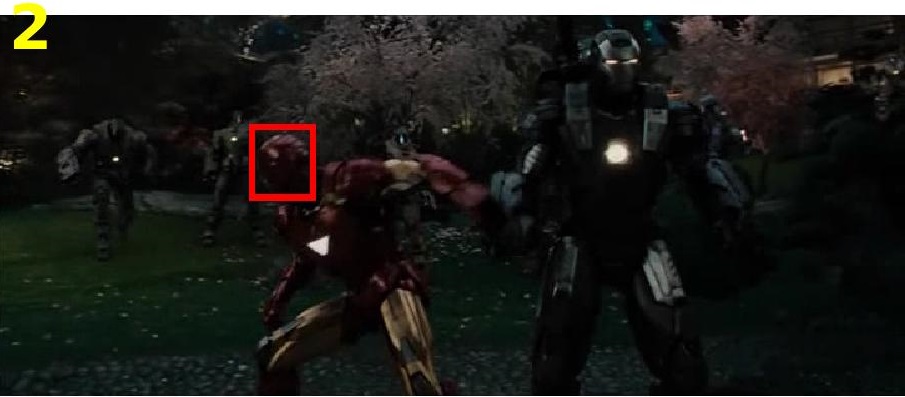}
	\includegraphics[width=0.24\linewidth]{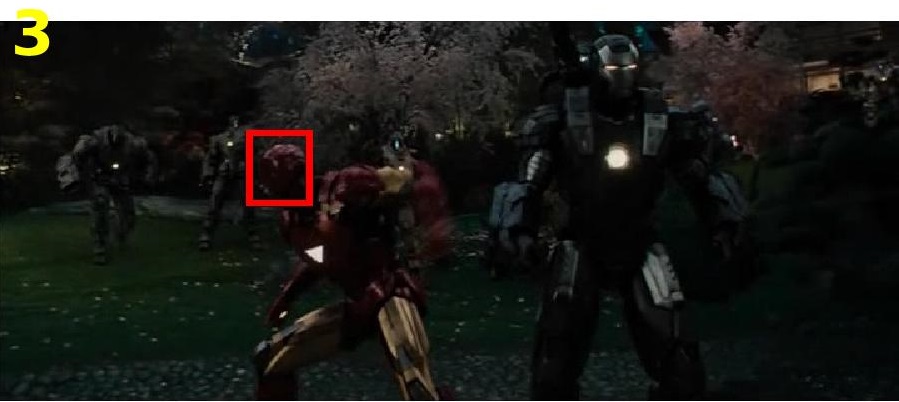}
	\includegraphics[width=0.24\linewidth]{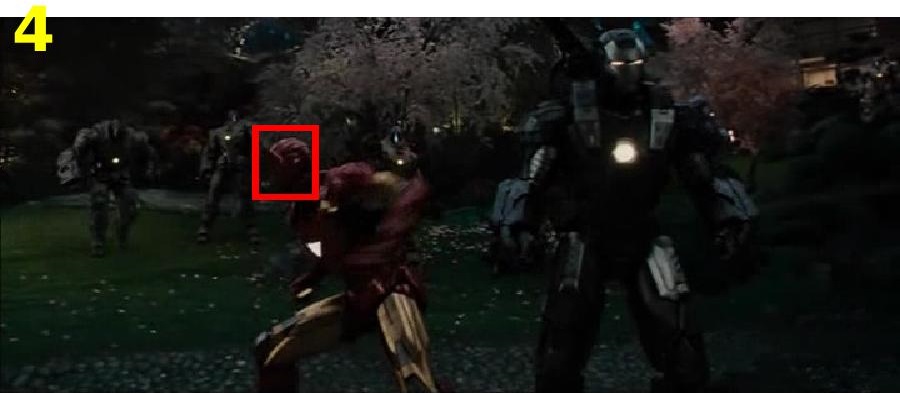}
	\includegraphics[width=0.24\linewidth]{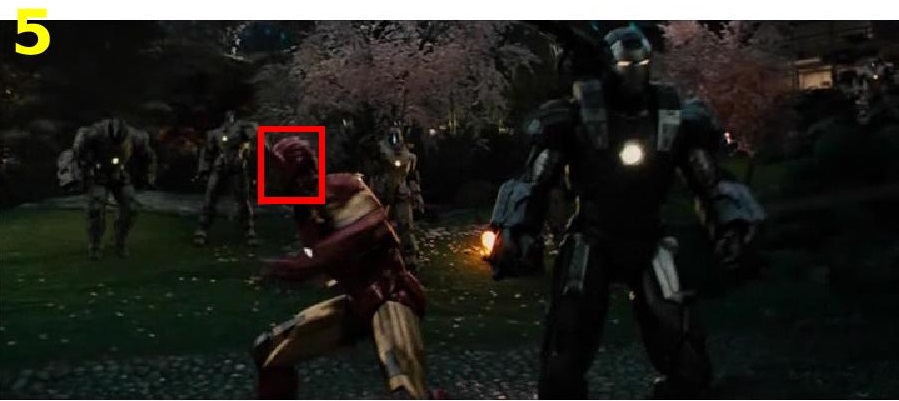}   
	
	\includegraphics[width=0.24\linewidth]{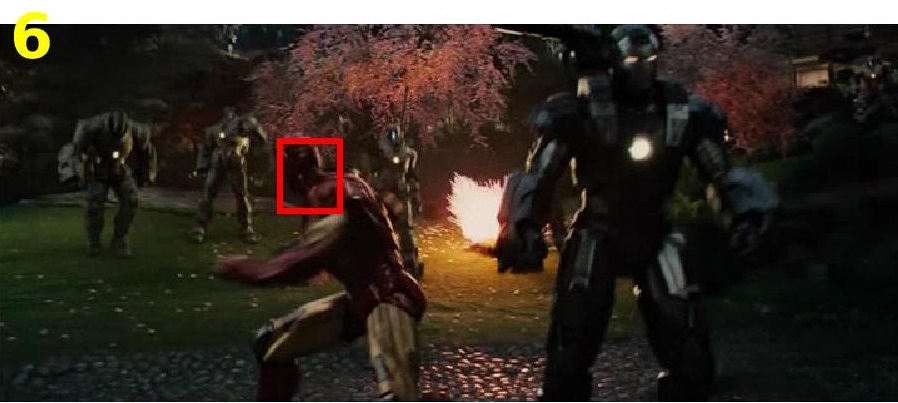}
	\includegraphics[width=0.24\linewidth]{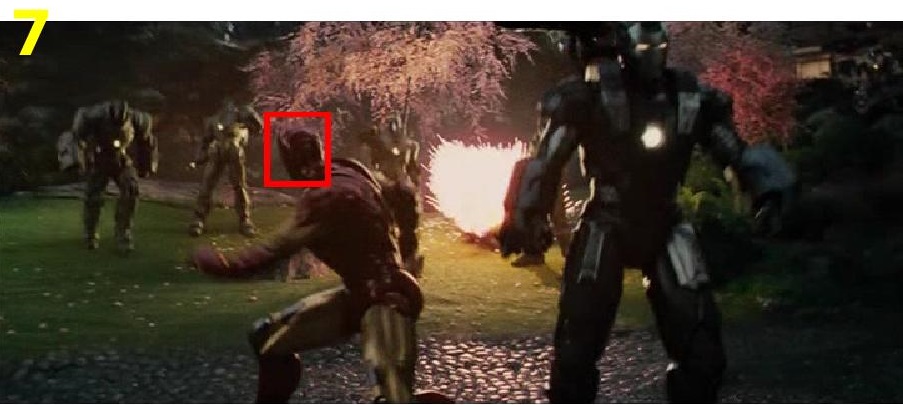}
	\includegraphics[width=0.24\linewidth]{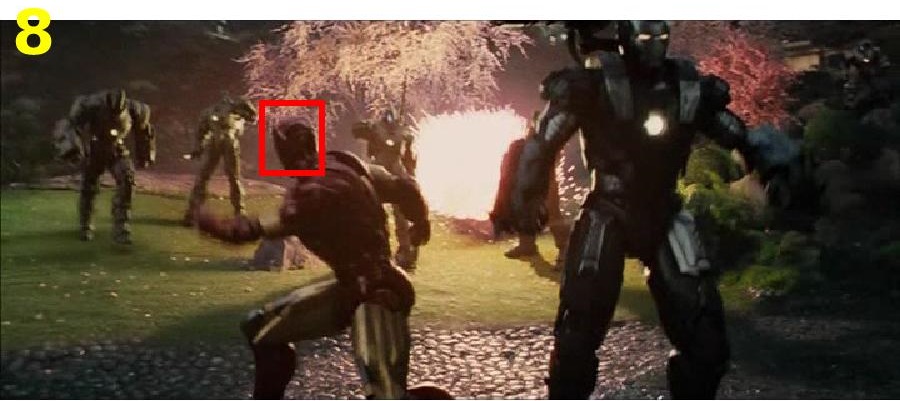}
	\includegraphics[width=0.24\linewidth]{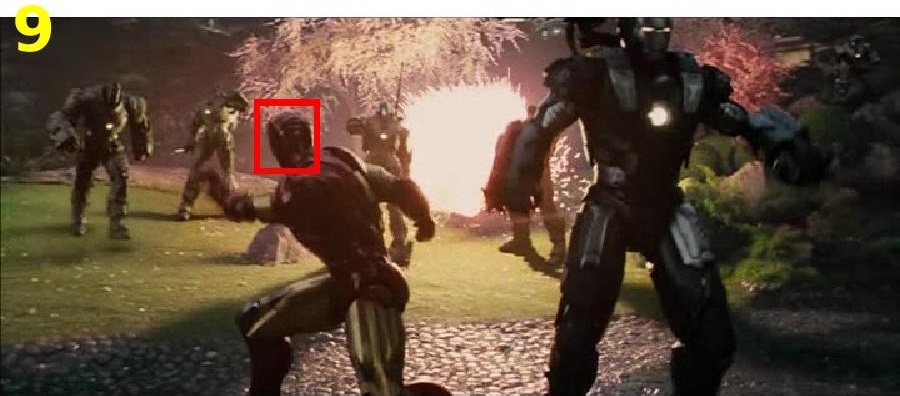}  
	
	\caption{frame 2-9 in the sequence of Ironman}
	\label{fig:fc_i}	
\end{figure}

After the proposed network has been trained, we delete the last fully-connected layer and add a fully-connected layer with two outputs, representing the target and the background, respectively. The parameters in the last layer are initialized randomly which subject to Gaussian distribution. The three fully-connected layers will be updated while tracking, but the convolution layers will not be changed.\\
In the beginning, the positive and negative examples from the first frame are used to fine-tune the network. Then, we select samples from the current frame near the target and send the samples to the network. The sample with the highest score will be obtained as the target in this frame. If the score is higher than a threshold $m$, both positive and negative samples near the target in the current frame are collected for updating. When the highest score is less than the threshold $m$, we use samples collected from the previous frames to update the fully-connected layers. $m$ is set to 0 in our experiments. \\
In MDNet, the network will be iterated 10 times if the highest score is less than the threshold $m$. But in our experiments, we find that a lot of iterations are useless and spend much time. So we define a loss threshold $l$. The iteration will stop if the loss of the network is less than $l$. It is a very simple strategy, but it is an effective one which will save a large amount of time. We use an example to explain this strategy and its consequence. The loss threshold $l$ is set to 0.01.\\
We use the sequence of Ironman in OTB100. When the score is less than $m$, which is 0, the network should be fine-tuned. When the score is larger than $m$, positive and negative samples will extracted from this frame for the next fine-tuning. When the network is being updated, it will iterated 10 times and will stop iteration if the loss is less than $l$, which is 0.01. We could see the results of frame 2-9 in Fig.~\ref{fig:fc_i} and know some parameters from Table.~\ref{fc_t}. In Table.~\ref{fc_t}, scores in frame 6-9 are less than the threshold $m$, so the network has to be updated. But the loss is very small in frame 6. The small loss means that the parameters in the network is been updated very well. In frame 7,8 and 9, we have to use the same positive and negative samples from frame 1,2,4 and 5(whose scores are larger than 0) to update the network, and because the loss is very small, these updates are ineffective and time consuming. So we use the loss threshold $l$ to control the update and it will stop the fine-tuning when the loss is less than $l$.\\

\begin{table}
	\centering
	\caption{the states of fine-tuning in the frame 2-9 of Ironman}
	\setlength{\tabcolsep}{1mm}
	\begin{tabular}{ccccccccc}   
		\toprule
		frame        & 2     & 3      & 4     & 5     & 6      & 7      & 8      & 9      \\  
		\midrule        
		score        & 0.392 & -1.443 & 3.199 & 1.444 & -5.294 & -3.340 & -2.419 & -1.221 \\ 
		loss         & -     & 0.03   & -     & -     &  0.006 & 0.004  & 0.003  & 0.001   \\ 
		iteration    & -     & 7      & -     & -     &  6     & 1      & 1      & 1      \\ 
		\bottomrule  
		\label{fc_t}
	\end{tabular}
\end{table}

We could see from the Fig.~\ref{fig:fc_i} that it can track the object very well even thought the scores in frame 6,7,8 and 9 are less than the threshold $m$. It means that $m$ is not suitable for this sequence. But there are 100 sequences in OTB100 and we can not change the threshold in each sequence. So we could just use the loss threshold $l$ to control the fine-tuning of the network.\\
The last row of Table.~\ref{fc_t} is the numbers of iteration which the network needs. We can see that it only needs $7+6+1+1+1=16$ iterations in these frames. But in MDNet, the network always has to be updated 10 times and the total number of iteration is $10\times5=50$. We know that updating the network when tracking will have huge influence on the speed of the tracker. By this method, we can reduce a large amount of time and it will not affect the tracking result. Because the iterations we delete are useless and ineffective.

Moreover, hard negative mining~\cite{32} and bounding box regression~\cite{25} have been employed. Experiments~\cite{26} show that both of them can improve the tracking performance. Hard negative mining is a trick to select effective samples. The negative samples will be sent to the network to get scores. And the samples with higher scores are taken as the hard negative samples. We use these hard negative samples instead of all the negative samples to update the fully-connected layers. The bounding box regression is just like what they used in R-CNN~\cite{11}. A simple linear regression model is trained to predict the target region using the features of the $3\textsuperscript{rd}$ convolution layer. The training of the bounding box regression is only conducted with the first frame. The learning rate is set to 0.0005 in the first frame and it is set to 0.0015 for online updating.\\


\section{Experiment}
We evaluated our network on a dataset: object tracking benchmark (OTB). Our code is implemented in MATLAB with matconvnet~\cite{35}. It runs at about 10 fps on TITAN Xp GPU.
\subsection{Experiment on OTB}
OTB~\cite{39} is a popular tracking benchmark. Each of those vedios is labeled with attributes. The evaluation is based on two metrics: center location and bounding box overlap ratio. Three evaluation methods have been employed: OPE(one pass evaluation), TRE(temporal robustness evaluation) and SRE(spatial robustness evaluation). The proposed method has been compared with state of the art methods including CSK~\cite{25}, Struct~\cite{12}, CF2~\cite{23} and Staple~\cite{2}. To train the network, 58 videos from VOT2013~\cite{47}, VOT2014~\cite{48} and VOT2015~\cite{49} have been employed, excluding the videos included in OTB100. The results of OTB100 are shown in Fig.~\ref{fig:OTB100}. And some examples of the visual tracking results are shown in Fig.~\ref{fig:OTBresults}. From these results it could be concluded that superior or comparable performance has been obtained via the proposed method.\\
\subsection{Comparisons of RoIAlign and RoIPool}
\begin{figure} 	
	\centering 	
	
	\includegraphics[width=0.45\linewidth]{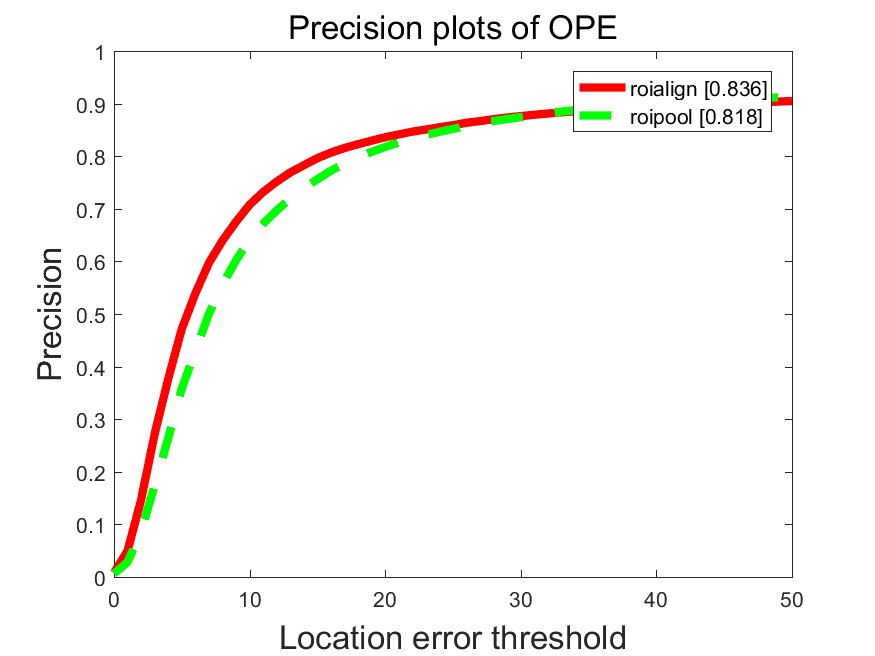}
	\includegraphics[width=0.45\linewidth]{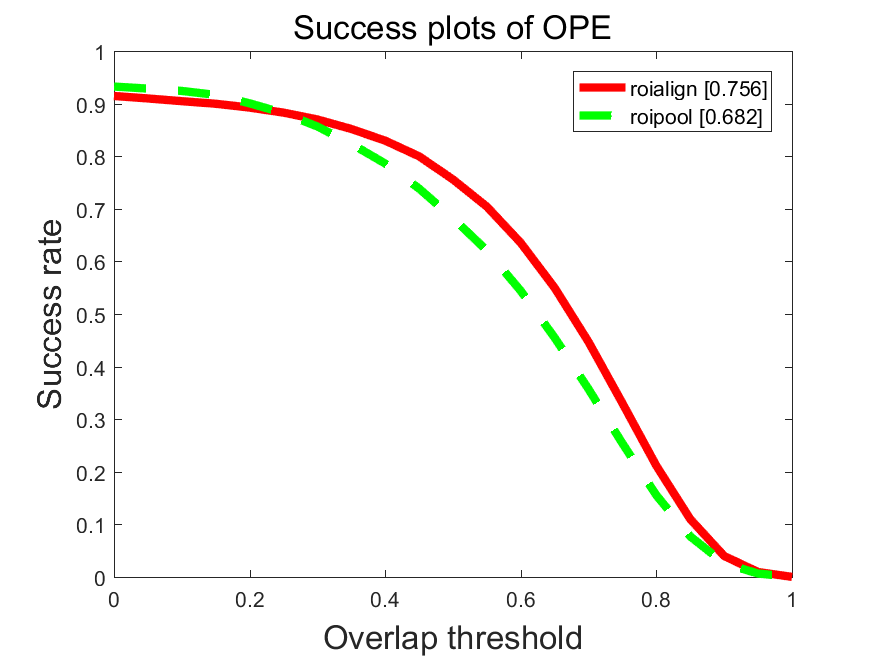}	
	\caption{The results of RoIPool and RoIAlign on OTB100}
	\label{fig:roipool}	
\end{figure}

Most of the networks using RoIPool from fast-RCNN to accelerate the convolution have obtained good performance on detection and classification task. However, for tracking tasks, the localization error will continue to be accumulated and enlarged along the working of the tracker. Even a small error of the current frame will affect the ones after it. So we use RoIAlign instead of RoIPool to improve the target localization precision. Fig.~\ref{fig:roipool}. shows the results of RoIAlign and RoIPool on OTB100.

\subsection{Comparisons of Feature selection network and simple trained network with 256 feature maps}
\begin{figure} 	
	\centering 	
	
	\includegraphics[width=0.45\linewidth]{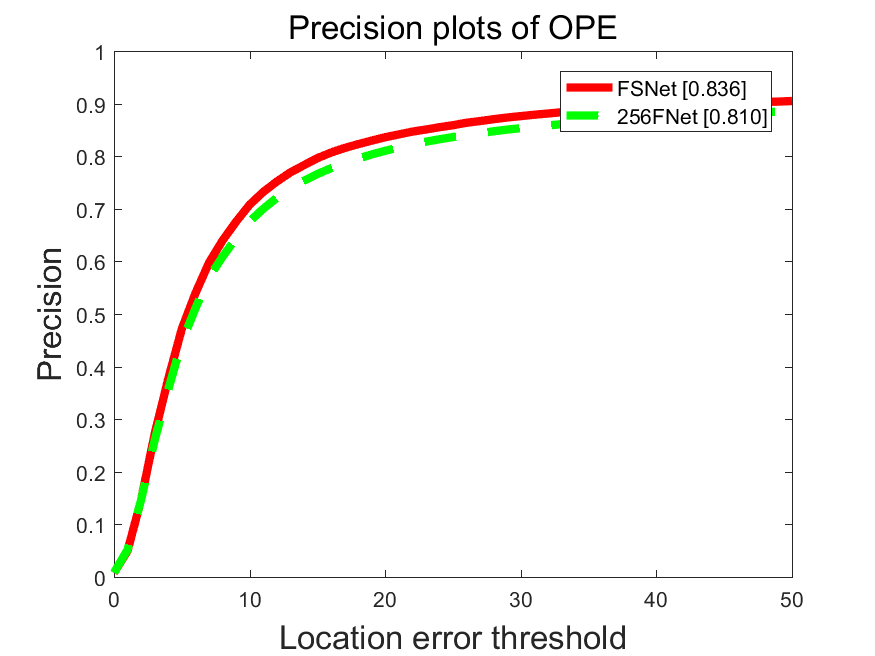}
	\includegraphics[width=0.45\linewidth]{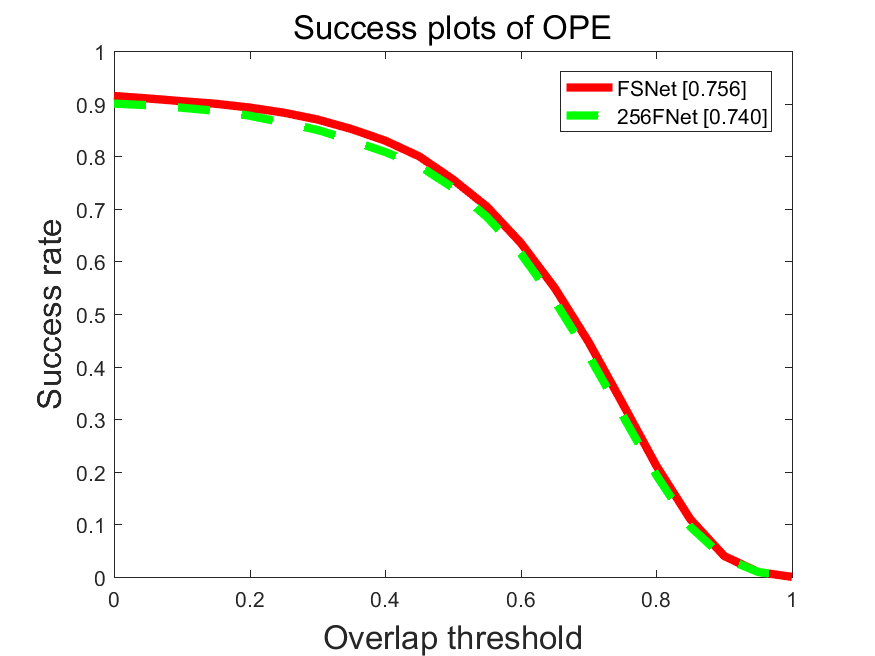}	
	\caption{The results of FSNet and 256FNet on OTB100}
	\label{fig:256FNet}	
\end{figure}

When training the network, there are 512 feature maps in the conv3. And before tracking, 256 feature maps are selected from the conv3. What if we just train a network with 256 feature maps in conv3( 256FNet). The result is in Fig.~\ref{fig:256FNet}. We can find that our feature selection network is better than a simple network which is trained with 256 feature.
\subsection{How many feature maps do we need?}
\begin{figure} 	
	\centering 	
	
	\includegraphics[width=0.45\linewidth]{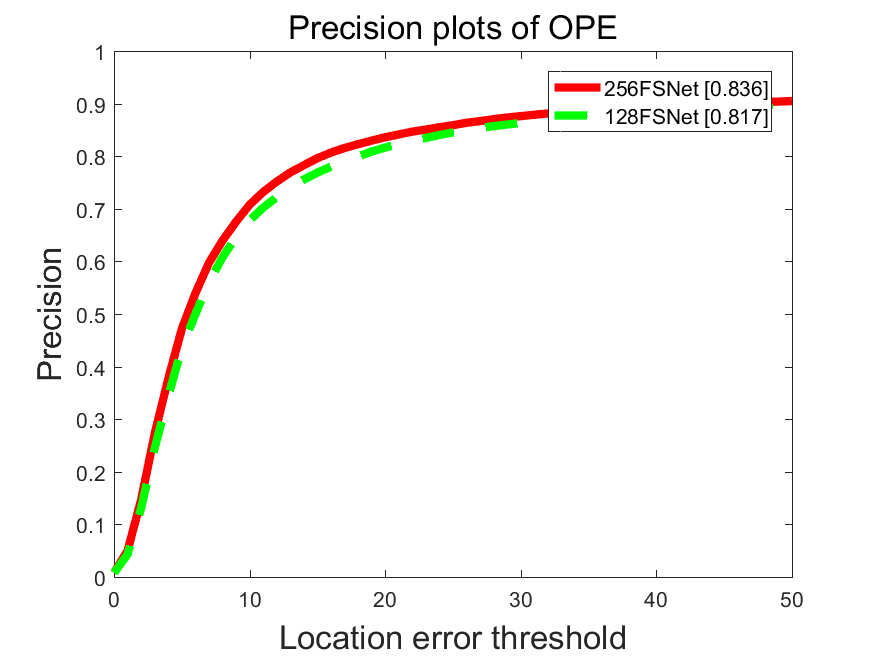}
	\includegraphics[width=0.45\linewidth]{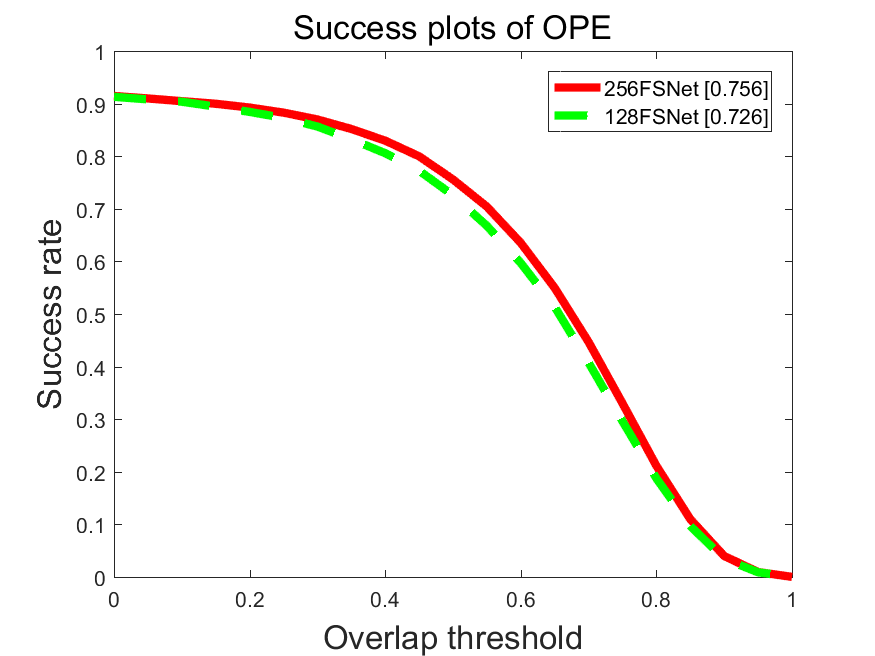}	
	\caption{The results of 256FSNet and 128FSNet on OTB100}
	\label{fig:128FSNet}	
\end{figure}

There are 512 feature maps in the original network. We select 256 feature maps and it has no influence on the precision. So we try to reduce the featues again and select 128 feature maps by mutual information. The results is shown in Fig.~\ref{fig:128FSNet}. It illustrate that the precision drops a little bit when 128 feature maps are selected. So 256 feature maps is a balance between the precision and computation.

\section{Conclusions}
A feature selection convolution neural network for visual tracking was proposed. To speed up the computation and maintain the accuracy in the meantime, a whole frame based convolution and RoIAlign has been employed. There are 3 convolution layers, 3 fully-connected layers and a RoIAlign layer between them. The RoIAlign layer will accelerate the convolution and performs better than RoIPool with higher target localization precision. The last fully-connected layer has $k$ branches which correspond to $k$ videos during training. Before tracking, useful feature maps will be selected by mutual information and others are abandoned. This strategy will reduce the complexity and computation significantly. During tracking, the last fully-connected layer is deleted and a new one which is randomly initialized is added. The first frame is used to train the fully-connected layers and update the parameters. A new strategy of fine-tuning the fully-connected layers is used to accelerate the update when tracking. Finally about 10 fps on GPU is reached and outstanding performance on a tracking benchmark, OTB, has been obtained.

\end{document}